\documentclass[10pt,twocolumn,letterpaper]{article}

\usepackage{cvpr}
\usepackage{times}
\usepackage{epsfig}
\usepackage{graphicx}
\usepackage{commath}
\usepackage{amsmath}
\usepackage{amssymb}
\usepackage{boldline}
\usepackage{float}
\usepackage{subfigure}
\usepackage{multirow}
\usepackage{cuted}
\usepackage{capt-of}

\usepackage[breaklinks=true,bookmarks=false]{hyperref}

\cvprfinalcopy 

\def\cvprPaperID{****} 

\ifcvprfinal\fi
\begin{document}

\title{SwapText: Image Based Texts Transfer in Scenes}


\author{Qiangpeng Yang, Hongsheng Jin, Jun Huang, Wei Lin\\
	Alibaba Group\\
	{\tt\small yqp0424@gmail.com,\{hongsheng.jhs, huangjun.hj, weilin.lw\}@alibaba-inc.com}
}

\maketitle
\thispagestyle{empty}

\begin{strip}\centering
	\includegraphics[width=0.87\textwidth]{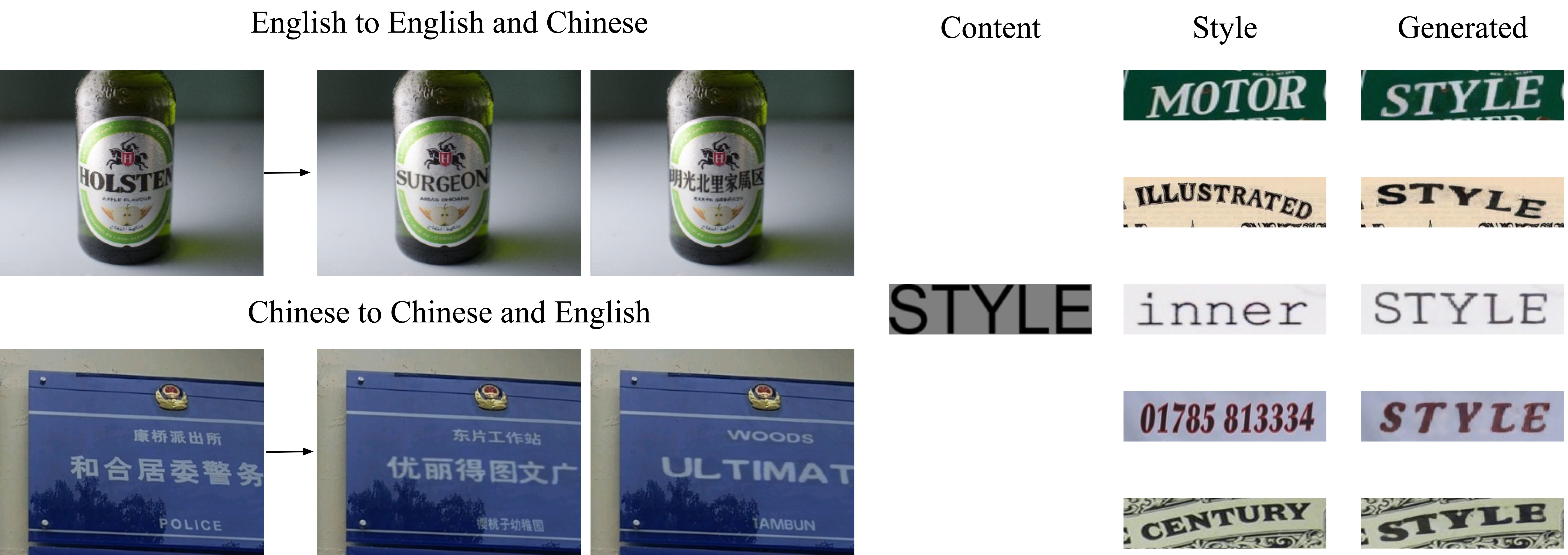}
	\captionof{figure}{Arbitrary text style transfer in scene text images. (\textit{Left}) Our model learns to perform diverse translation between multilanguage. (\textit{Right}) Style-guide transfer
		\label{fig:intro}}
\end{strip}

\begin{abstract}
   Swapping text in scene images while preserving original fonts, colors, sizes and background textures is a challenging task due to the complex interplay between different factors.
   In this work, we present SwapText, a three-stage framework to transfer texts across scene images.
   First, a novel text swapping network is proposed to replace text labels only in the foreground image. 
   Second, a background completion network is learned to reconstruct background images.
   Finally, the generated foreground image and background image are used to generate the word image by the fusion network.
   Using the proposing framework, we can manipulate the texts of the input images even with severe geometric distortion.
   Qualitative and quantitative results are presented on several scene text datasets, including regular and irregular text datasets.
   We conducted extensive experiments to prove the usefulness of our method such as image based text translation, text image synthesis, \etc.
\end{abstract}

\section{Introduction}

Imagine being able to swap text in scene images while keeping the original fonts, colors, sizes and background textures within seconds, and without hours of image editing.
In this work, we aim to realize this goal with an algorithm that automatically replaces the text in scene images.
The core challenge of text swapping lies in generating visually realistic text and keeping coherent style with the original text.


Text swapping or text replacement is relevant in many scenarios including text detection, text recognition, text transfer in posters and other creative applications.
For text detection and recognition tasks, text swapping is a very useful data augmentation approach.
Witness the great success of deep neural networks (DNN) in various computer vision tasks, obtaining large amounts of annotated training images has become the bottleneck for training DNN models. 
The easiest and most widely used methods augment training images by geometric transformation, such as translation, rotation and flipping, \etc.
Recently, image synthesis based approaches~\cite{jaderberg2014synthetic, gupta2016synthetic, zhan2018verisimilar} have been proposed for training text detection and recognition models.
These approaches create new images from text-free images by modeling the physical behaviors of light and energy in combination of different rendering techniques. 
However, the synthetic images do not fully cohere with the images in scenes, which is critically important while applying the synthesized images to train DNN models.

In most recent years, many image generation models, such as generative adversarial networks (GANs)~\cite{goodfellow2014generative} , variational autoencoders (VAE)~\cite{kingma2013auto}, and autogressive models~\cite{oord2016pixel} have provided powerful tools for realistic image generation tasks.
In~\cite{iizuka2017globally, yu2018generative, wang2018image}, GANs are used for image completion that generates visually realistic and semantically plausible pixels for the missing regions. \cite{ma2017pose, han2018viton, raj2018swapnet, ma2018disentangled} have exploited these networks to generate novel person  images with different poses or garments. 

Based on GANs, we present a unified framework SwapText for text swapping in scenes. A few examples can be seen in Figure~\ref{fig:intro}.
We adopt a divide-and-conquer strategy, decompose the problem into three sub-networks, namely text swapping network, background completion network and the fusion network.
In the text swapping network, the features of content image and style image are extracted simultaneously and then combined by a self-attention network.
To better learn the representation of content image, we use a Content Shape Transformation Network (CSTN) to transform the content image according to the geometrical attributes of the style image.
According to our experiments, this transformation process has significantly improved image generation, especially for perspective and curved images.
Then, a background completion network is used to generate the background image of style image.
Because we need to erase the original text stroke pixels in the style image and fill with appropriate texture according to the content image.
Finally, the output of text swapping network and background completion network are fed into the fusion network to generate more realistic and semantically coherent images.
The whole framework are end-to-end trainable, and extensive experiments on several public benchmarks demonstrate its superiority in terms of both effectiveness and efficiency.

Our contributions are summarized as follows:
\begin{itemize}
	\item We design an end-to-end framework namely SwapText, which contains three sub-networks, text swapping network, background completion network and the fusion network.
	\item We propose a novel text swapping network that replace the text in the scene text images, while keeping the original style. 
	\item We demonstrate the effectiveness of our method for scene text swapping with high-quality visual results, and also show its application to text image synthesis, image based text translation, \etc.
\end{itemize}

\section{Related Work}

\paragraph{Text Image Synthesis}
Image synthesis has been studied extensively in computer graphics research~\cite{debevec2008rendering}.
Text image synthesis is investigated as a data augmentation approach for training accurate and robust DNN models.
For example, Jaderberg \etal~\cite{jaderberg2014synthetic} use a word generator to generate synthetic word images for text recognition task.
Gupta \etal~\cite{gupta2016synthetic} develop a robust engine to generate synthetic text image for both text detection and recognition tasks. 
The target of text image synthesis is to insert texts at semantically sensible regions within the background image.
Many factors affect the true likeness of the synthesized text images, such as text size, text perspective, environment lighting, \etc.
In~\cite{zhan2018verisimilar}, Zhan \etal achieve verisimilar text image synthesis by combining three designs including semantic coherence, visual attention, and adaptive text appearance.
Although the text images synthesis are visually realistic, there are many differences between synthetic images and real images.
For instance, comparing to the real images the fonts of text and background image in synthetic images are very limited.

In most recently, GAN based image synthesis technology has been further explored.
In ~\cite{zhan2018spatial}, Zhan \etal present an spatial fusion GAN that combines a geometry synthesizer and an appearance synthesizer to achieve synthesis realism in both geometry and appearance spaces.
Yang \etal~\cite{Yang:2019ud} use bidirectional shape matching framework control the crucial stylistic degree of the glyph through an adjustable parameter.
GA-DAN~\cite{zhan2019ga} present an interesting work that is capable of modelling cross domain shifts concurrently in both geometry space and appearance space.
In~\cite{azadi2018multi}, MC-GAN is proposed for font style transfer in the set of letters from A to Z.
Wu \etal~\cite{Wu:2019ww} propose an end-to-end trainable style retention network to edit text in natural images. 

\paragraph{Image Generation}
With the great success of generative models, such as GANs~\cite{goodfellow2014generative}, VAEs~\cite{kingma2013auto} and autogressive models~\cite{oord2016pixel}, realistic and sharp image generation has attracted more and more attention lately.
Traditional generative models use GANs~\cite{goodfellow2014generative} or VAEs~\cite{kingma2013auto} to map a distribution generated by noise \textit{z} to the distribution of real data.
For example, GANs~\cite{goodfellow2014generative} are used to generate realistic faces~\cite{yin2017towards, berthelot2017began, karras2018progressive} and birds~\cite{reed2016generative}.

To control the generated results, Mirza \etal~\cite{mirza2014conditional} proposed conditional GANs.
They generate MNIST digits conditioned on class labels.
In~\cite{karacan2016learning}, karacan \etal generate realistic outdoor scene images based on the semantic layout and scene attributes, such as day-night, sunny-foggy.
Lassner \etal~\cite{lassner2017a} generated full-body images of persons in clothing based on fine-grained body and clothing segments.
The full model can be conditioned on pose, shape, or color.
Ma \etal~\cite{ma2017pose, ma2018disentangled} generate person images based on images and poses.
Fast face-swap is proposed in~\cite{korshunova2017fast} to transform an input identity to a target identity while preserving pose, facial expression and lighting.

\begin{figure*}[th]
	\centering
	\includegraphics[width=0.9\textwidth]{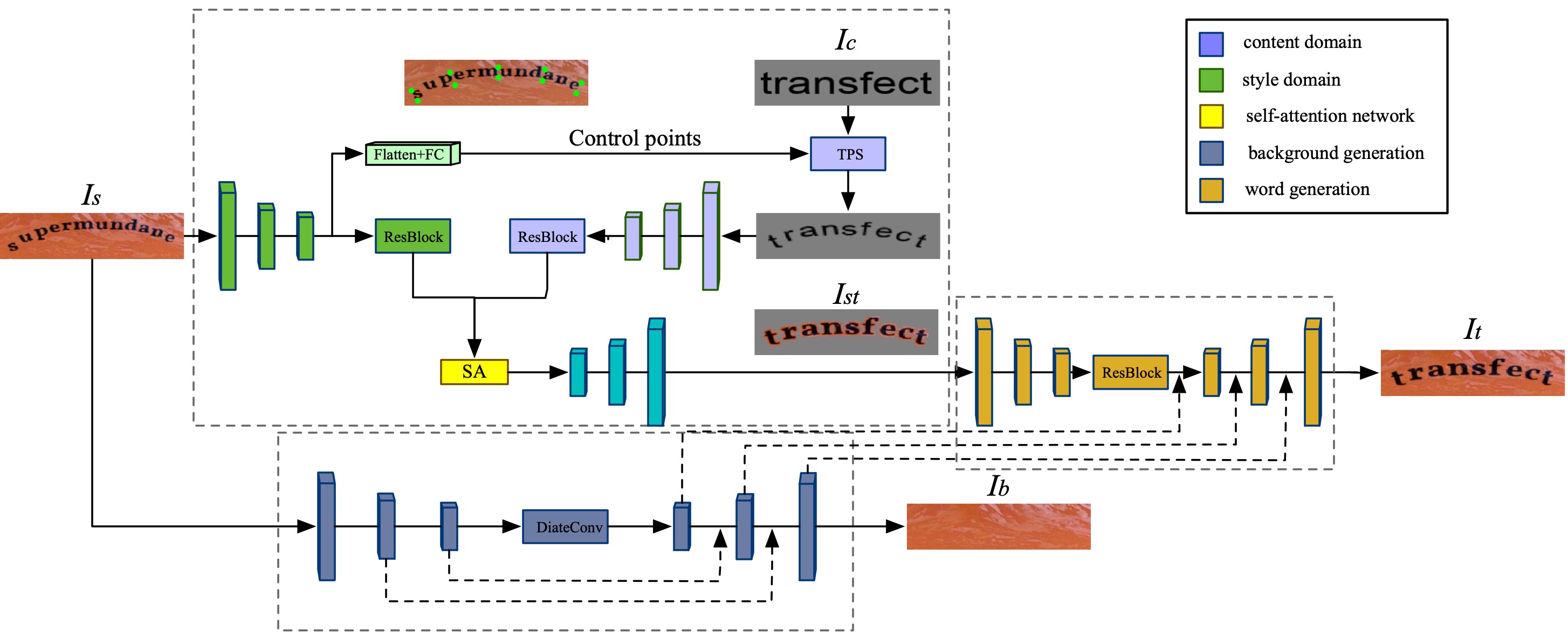}
	\caption{The framework of our proposed method. It contains three sub-networks: text swapping network, background completion and fusion network.}
	\label{fig:framework}
\end{figure*}

\paragraph{Image Completion}
Recently, GAN-based approaches have emerged as a promising paradigm for image completion.
Iizuka \etal~\cite{iizuka2017globally} proposed to use global and local discriminators as adversarial losses, where both global and local consistency are enforced.
Yu \etal~\cite{yu2018generative} use a contextual attention layer to explicitly attend on related feature patches at distant spatial locations.
Wang \etal~\cite{wang2018image} use a multi-column network to generate different image component in a parallel manner, and an implicit diversified MRF regularization is adopted to enhance local details.

\section{Methodology}

Given a scene text image $I_{s}\in \mathbb{R}^{H\times W\times3}$, our goal is to replace the text based on a content Image  $I_{c}\in \mathbb{R}^{H\times W\times3}$ while keeping the original style.
As illustrated in Figure~\ref{fig:framework}, our framework consists of text swapping network, background completion network and the fusion network.
The text swapping network firstly extracts the style features from $I_{s}$ and content features from $I_{c}$, then combine these two features by a self-attention network.
To learn a better representation of content, we use a Content Shape Transformation Network (CSTN) to transform the content image $I_{c}$ according to the geometrical attributes of the style image  $I_{s}$.
The background completion network is used to reconstruct the original background images $I_{b}$ of style image $I_{s}$.
Finally, the outputs of text swapping network and background completion network are fused by the fusion network to generate the final text images.

\subsection{Text Swapping Network}

Text instances in real-world scenarios have diverse shapes, \eg, in horizontal, oriented or curved forms.
The main purpose of text swapping network is to replace the content of the style image $I_{s}$, while keeping the original style, especially text shapes.
To improve the performance of irregular text image generation, we propose a Content Shape Transformation Network (CSTN) to map the content image into the same geometric shape of the style image.
Then the style image and transformed content image are encoded by 3 down-sampling convolutional layers and several residual blocks.
To combine the style and content features adequately, we feed them into a self-attention network.
For decoding, 3 up-sampling deconvolutional layers are used to generate the foreground images $I_{f}$.

\subsubsection{Content Shape Transformation Network}

\begin{figure}[th]
	\centering
	\includegraphics[width=0.45\textwidth]{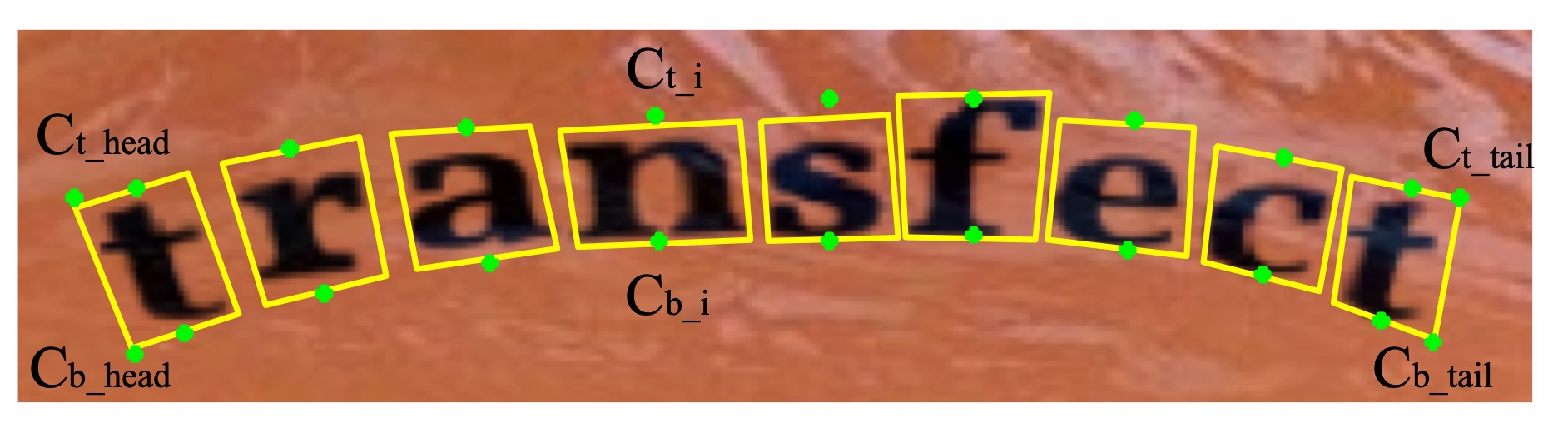}
	\caption{Illustration of text shape definition.}
	\label{fig:geo_define}
\end{figure}

The definition of text shape is critical for content shape transformation.
Inspired by the text shape definition in text detection~\cite{long2018textsnake:} and text recognition~\cite{Yang:2019uk} field, the geometrical  attributes of text can be defined with $2K$ fiducial points $P = \{p_{1}, p_{2}, ..., p_{2K}\}$, which is illustrated in Figure~\ref{fig:geo_define}.

A text instance can be viewed as an ordered character sequence $T = \{C_{1}, ..., C_{i}, ..., C_{n}\}$, where $n$ is the number of characters.
Each character $C_{i}$ has a bounding box $B_{i}$, which is annotated with a free-form quadrilateral.
First, we construct two center point lists $C_{top} =  \{C_{t\_head}, C_{t\_1}, ..., C_{t\_n}, C_{t\_tail}\}$ and $C_{bottom} =  \{C_{b\_head}, C_{b\_1}, ..., C_{b\_n}, C_{b\_tail}\}$, which contains the top center and bottom center for each $B_{i}$.
Then we evenly spaced sampling $K$ fiducial points in $C_{top}$ and $C_{bottom}$.
For the points not in $C_{top}$ or $C_{bottom}$, the values are linearly interpolated with two nearest center points.
In this way, the shape of the text instance is precisely described by the fiducial points.
In our experiments, $K$ is set to 5.

To yield the text shape of input style image, we employ a lightweight predictor which shares the down-sampling convolutional layers with style image encoder, as illustrated in Figure~\ref{fig:framework}.
The output of this predictor is $\hat{P} = \{\hat{p}_{1}, \hat{p}_{2}, ... \hat{p}_{2K}\}$, which represents the geometrical attributes of the input image.
We adopt $smooth_{L_{1}}$ loss the loss function of this predictor, 
\begin{equation}
\mathcal{L}_{P}=\frac{1}{2K}\sum_{i=1}^{2K}smooth_{L_{1}}(p_i - \hat{p}_i),
\end{equation}

\begin{figure}[th]
	\centering
	\includegraphics[width=0.45\textwidth]{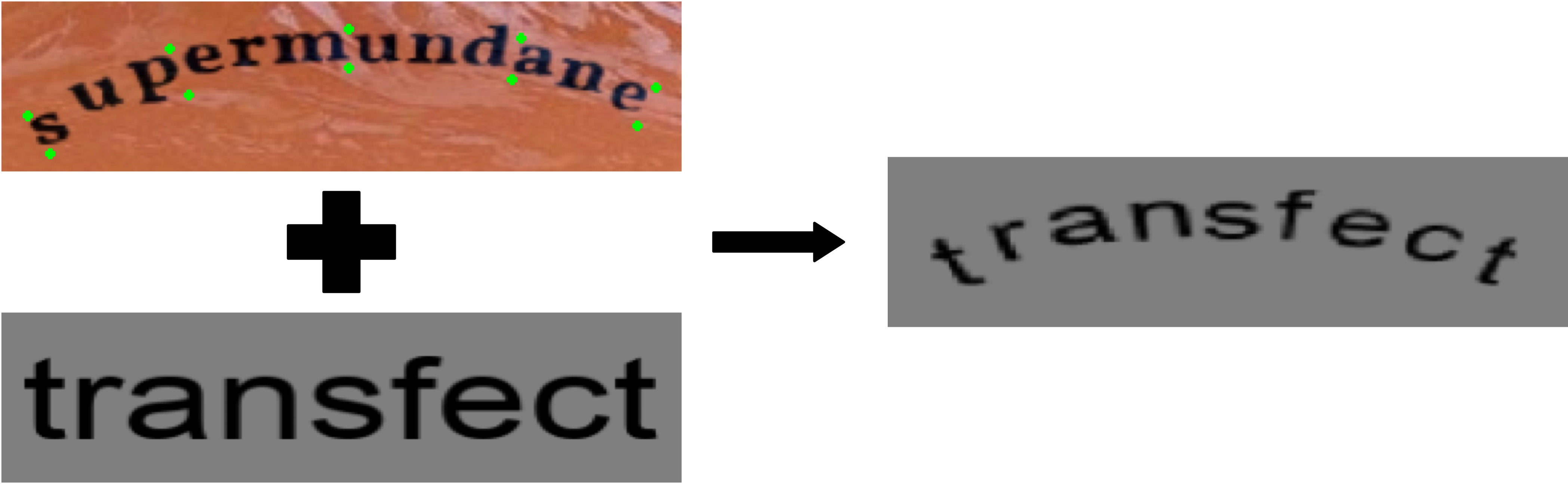}
	\caption{The shape transform process of content image.}
	\label{fig:shape_transform}
\end{figure}

Given the geometrical attributes of style image, we transform the content image through the Thin-Plate-Spline (TPS) module.
The transform process is shown in Figure~\ref{fig:shape_transform}.

\subsubsection{Self-attention Network}

\begin{figure}[th]
	\centering
	\subfigure[Self-attention network.]{
		\includegraphics[width=0.45\textwidth]{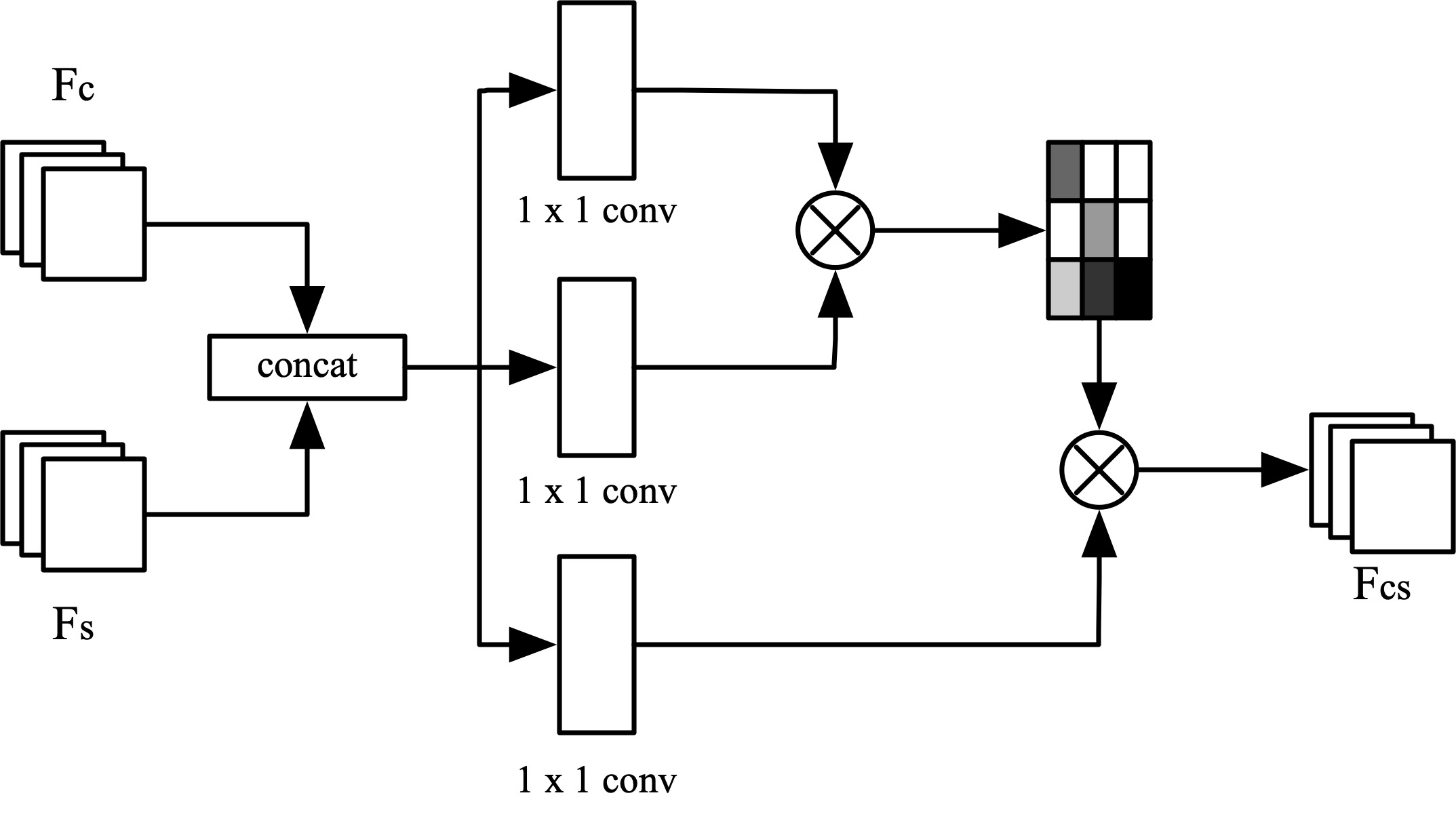}
	}
	\centering
	\subfigure[Multi-level stylization.]{
		\includegraphics[width=0.45\textwidth]{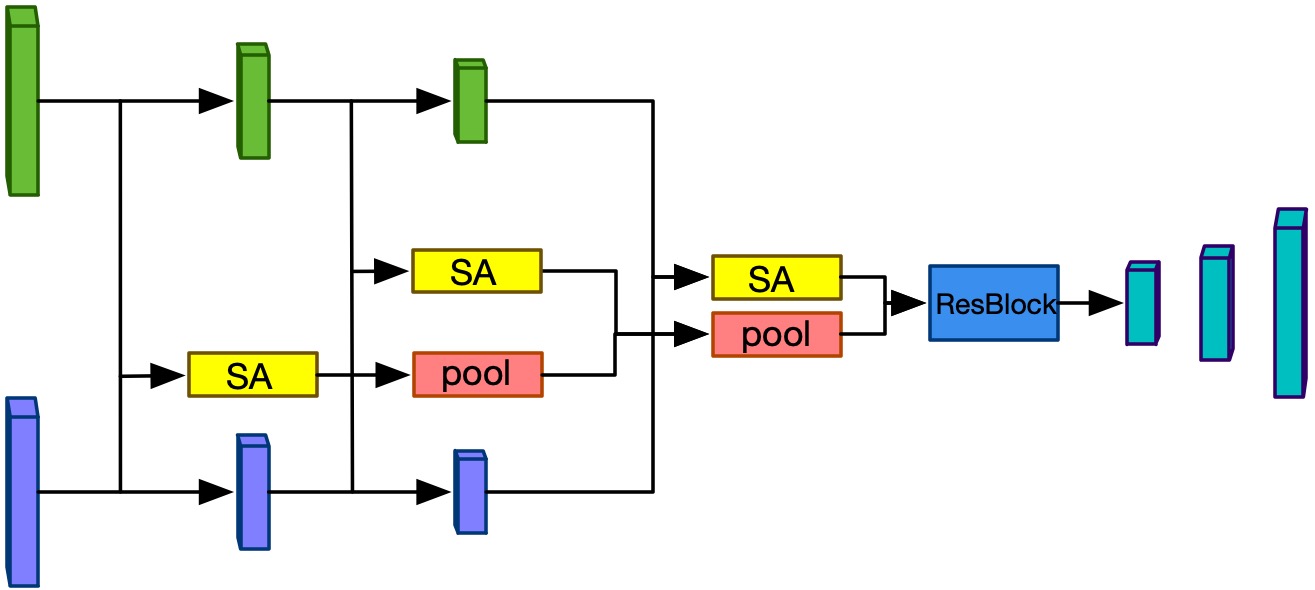}
	}
	
	\caption{Architecture of self-attention network. (a) Self-attention network. (b) Multi-level self-attention.}
	\label{fig:sanet}
\end{figure}

After encoding the content and style images, we feed both feature maps to a self-attention network that automatically learns the correspondences between the content feature map $F_{c}$ and style feature map $F_{s}$.
The output feature map is $F_{cs}$, and the architect of self-attention network is presented in Figure~\ref{fig:sanet} (a)

The content feature $F_{c}$ and style feature $F_{s}$ are firstly concatenated along their depth axis.
Then we follow the similar self-attention mechanism in~\cite{zhang2018self-attention} to produce the output feature map $F_{cs}$.

We adopt the $L_{1}$ loss to as our text swapping network loss function, which is as follows,
\begin{equation}\label{loss_fg_G1}
\mathcal{L}_{swap} = \norm[1]{G_{swap}(I_{s}, I_{t}) - I_{st}}_1,
\end{equation}
where $G_{swap}$ denotes the text swapping network, and $I_{st}$ is the ground truth of text swapping network.

In addition to this single-level stylization, we further develop a multi-level stylization pipeline, as depicted in Figure~\ref{fig:sanet} (b).
We apply self-attention network sequentially to multiple feature layers to generate more realistic images.

\subsection{Background Completion Network}

The text swapping network mainly focus on the foreground image generation, while the background images also play a important role of the final image generation.
To generate more realistic word image, we use a background completion network to reconstruct the background image, whose architecture is illustrated in Table~\ref{table:bg_completion}.
Most existing image completion approaches fill the pixels of the image by borrowing or copying textures from surrounding regions.
The general architecture follows an encoder-decoder structure, we use dilated convolutional layer after the encoder to compute the output pixel with larger input area,
By using dilated convolutions at lower resolutions, the model can effectively "see" a larger area of the input image.

\begin{table}[th]
	\centering
	\fontsize{8}{8}\selectfont
	\setlength{\tabcolsep}{10pt}
	\renewcommand{\arraystretch}{2.0}
	\caption{Architecture of background completion network.}
	\begin{tabular}{ccccc}
		\hlineB{3.5}
		Type & Kernel & Dilation & Stride & Channels \\ \hline
		conv & $5 \times 5$ & 1 & $1 \times 1$ & 32 \\  \hline
		conv & $3 \times 3 $ & 1 & $2 \times 2 $ & 64 \\
		conv & $3 \times 3 $ & 1 & $1 \times 1 $ & 64 \\
		conv & $3 \times 3 $ & 1 & $1 \times 1 $ & 64 \\ \hline
		conv & $3 \times 3 $ & 1 & $2 \times 2 $ & 128 \\
		conv & $3 \times 3 $ & 1 & $1 \times 1 $ & 128 \\
		conv & $3 \times 3 $ & 1 & $1 \times 1 $ & 128 \\ \hline
		conv & $3 \times 3 $ & 1 & $2 \times 2 $ & 256 \\
		conv & $3 \times 3 $ & 1 & $1 \times 1 $ & 256 \\
		conv & $3 \times 3 $ & 1 & $1 \times 1 $ & 256 \\ \hline
		diated conv & $3 \times 3 $ & 2 & $1 \times 1 $ & 256 \\
		diated conv & $3 \times 3 $ & 4 & $1 \times 1 $ & 256 \\
		diated conv & $3 \times 3 $ & 8 & $1 \times 1 $ & 256 \\ \hline
		deconv & $3 \times 3 $ & 1 & $\frac{1}{2} \times \frac{1}{2} $ & 256 \\
		conv & $3 \times 3 $ & 1 & $1 \times 1 $ & 256 \\
		conv & $3 \times 3 $ & 1 & $1 \times 1 $ & 256 \\ \hline
		deconv & $3 \times 3 $ & 1 & $\frac{1}{2} \times \frac{1}{2} $ & 128 \\
		conv & $3 \times 3 $ & 1 & $1 \times 1 $ & 128 \\
		conv & $3 \times 3 $ & 1 & $1 \times 1 $ & 128 \\ \hline
		deconv & $3 \times 3 $ & 1 & $\frac{1}{2} \times \frac{1}{2} $ & 64 \\
		conv & $3 \times 3 $ & 1 & $1 \times 1 $ & 64 \\
		conv & $3 \times 3 $ & 1 & $1 \times 1 $ & 64 \\ \hline
		output & $3 \times 3 $ & 1 & $1 \times 1 $ & 3  \\
		
		\hlineB{3.5}
	\end{tabular}
	
	\label{table:bg_completion}
\end{table}

The background completion network is optimized with both $\mathcal{L}_{1}$ loss and GAN loss. 
We use $G_{b}$ and $D_{b}$ to denote the background generator and discriminator, the overall loss for background generation are as follows,
\begin{equation}\label{loss_b}
\begin{aligned}
\mathcal{L}_{B} =  \mathbb{E}[\log D_{b}(I_{b}, I_{s}) &  +\log(1 - D_{b}(\hat{I}_{b}, I_{s}))]  + \\
							&\lambda_{b}\norm[1]{I_{b} -\hat{I}_{b}}_1,
\end{aligned}
\end{equation}
where $I_b$ and $\hat{I}_{b}$ are ground truth and predicted background images. $\lambda_{b}$ is the balance factor and is set to $10$ in our experiments.

\subsection{Fusion Network}

In this stage, the output of text swapping network and background completion network are fused to generate the complete text images.
As the pipeline illustrated in Figure~\ref{fig:framework}, the fusion network follow an encoder-decoder architecture.
Similar to ~\cite{Wu:2019ww}, we connect the decoding feature maps of the background completion network to the corresponding feature maps with the same resolution in the up-sampling phase of the fusion decoder.
We use $G_{fuse}$ and $D_{fuse}$ to denote the generator and discriminator network respectively.
The loss function of fusion network can be formulated as follows,
\begin{equation}\label{loss_f}
\begin{aligned}
\mathcal{L}_{F} =  \mathbb{E}[\log D_{fuse}(I_{t}, I_{c}) &  +\log(1 - D_{fuse}(\hat{I}_{t}, I_{c}))]  + \\
&\lambda_{fuse}\norm[1]{I_{t} -\hat{I}_{t}}_1,
\end{aligned}
\end{equation}
where $\hat{I}_{t}$ is the output of the generator and $\lambda_{fuse}$ is the balance factor which is set to $10$ in our experiment.

In order to make more realistic images, we also introduce VGG-loss to the fusion module following the similar idea of style transfer network~\cite{gatys2016image, Park:2018uq}.
There two parts of VGG-loss, the perceptual loss and style loss, as follows,
\begin{equation}\label{loss_vgg}
\begin{aligned}
\mathcal{L}_{vgg} &= \lambda_{1} \mathcal{L}_{per} + \lambda_{2} \mathcal{L}_{style} \\
\mathcal{L}_{per} &= \mathbb{E}[\sum_{i}  \norm[1]{ \phi_{i}(I_{t}) - \phi_{i}(\hat{I}_{t}) }_1] \\
\mathcal{L}_{style} &= \mathbb{E}_{j}[\norm[1]{ G^{\phi}_{j}(I_{t}) - G^{\phi}_{j}(\hat{I}_{t}) }_1],
\end{aligned}
\end{equation}
where $\phi_{i}$ is the activation map from $relu1\_1$ to $relu5\_1$ layer of VGG-19 model.
G is the Gram matrix. $\lambda_{1}$ and $\lambda_{2}$ are the balance factors respectively.

The loss function of the whole framework is:
\begin{equation}\label{loss_final}
\mathcal{L} = \mathcal{L}_{P} + \mathcal{L}_{swap} + \mathcal{L}_{B} + \mathcal{L}_{F} + \mathcal{L}_{vgg} 
\end{equation}

\section{Experiments}

\subsection{Implementation Details}

We follow the similar idea in~\cite{Wu:2019ww} to generate pairwised synthetic images with same style.
We use over $1500$ fonts and $10000$ background images to generate a total of $1$ million training images and $10000$ test images.
The input images are resized to $64 \times 256$ and the batch size is $32$.
All weights are initialized from a zero-mean normal distribution with a standard deviation of $0.01$
The Adam optimizer~\cite{kingma2014adam} with $\beta_{1} = 0.9$ and $\beta_{2} = 0.999$ is used to optimize the whole framework.
The learning rate is set to $0.0001$ in the training phase.
We implement our model under the TensorFlow framework~\cite{abadi2016tensorflow}.
Most modules of our method are GPU-accelerated.

\subsection{Benchmark Datasets}

We evaluate our proposed method on several public benchmark datasets.

\noindent \textbf{IIIT 5K-Words}~\cite{mishra2012scene} (IIIT5K) contains $3000$ cropped word images for testing, while each image is assigned with a 50-word lexicon and a 1k-word lexicon.
All images are collected from the Internet.

\noindent \textbf{Street View Text}~\cite{wang2011end} (SVT) is collected from Google Street View, which contains $647$ images in the test set. Many images are severely corrupted by noise and blur, or have very low resolutions. Each image is associated with a 50-word lexicon.

\noindent \textbf{ICDAR 2013}~\cite{karatzas2013icdar} (IC13) is obtained from the Robust Reading Chaallenges 2013.
We follow the protocol proposed by~\cite{wang2011end}, where images contain non-alphanumeric characters or those having less than three characters are not taken into consideration.
After filtering samples, the dataset contains $857$ images without any pre-defined lexicon.

\noindent \textbf{ICDAR 2015}~\cite{karatzas2015icdar} (IC15) is more challenging than IC13, because most of the word images suffer from motion blur and low resolution. Moreover, many images contain severe geometric distortion, such as arbitrary oriented, perspective or curved texts.
We filter images following the same protocol in IC13.

\noindent \textbf{SVT-Perspective}~\cite{quy2013recognizing} (SVTP) contains 639 cropped images for testing, which are collected from side-view angle snapshots in Google Street View. Most of the images in SVT-Perspective are heavily deformed by perspective distortion.

\noindent \textbf{CUTE80}~\cite{risnumawan2014robust} is collected for evaluating curved text recognition. It contains 288 cropped images for testing, which is selected from 80 high-resolution images taken in the natural scene.

\subsection{Evaluation Metrics}
We adopt the commonly used metrics in image generation to evaluate our method, which includes the following:
\begin{itemize}
	\item MSE, also known as l2 error.
	\item PSNR, which computes the the ratio of peak signal to noise.
	\item SSIM, which computes the mean structural similarity index between two images
	\item Text Recognition Accuracy, we use text recognition model CRNN~\cite{shi2017end} to evaluate generated images.
	\item Text Detection Accuracy, we use text detection model EAST~\cite{zhou2017east} to evaluate generated images.
\end{itemize}
A lower l2 error or higher SSIM and PSNR mean the results are similar to ground truth. 

\subsection{Ablation Studies}
\begin{figure}[th]
	\centering
	\includegraphics[width=0.48\textwidth]{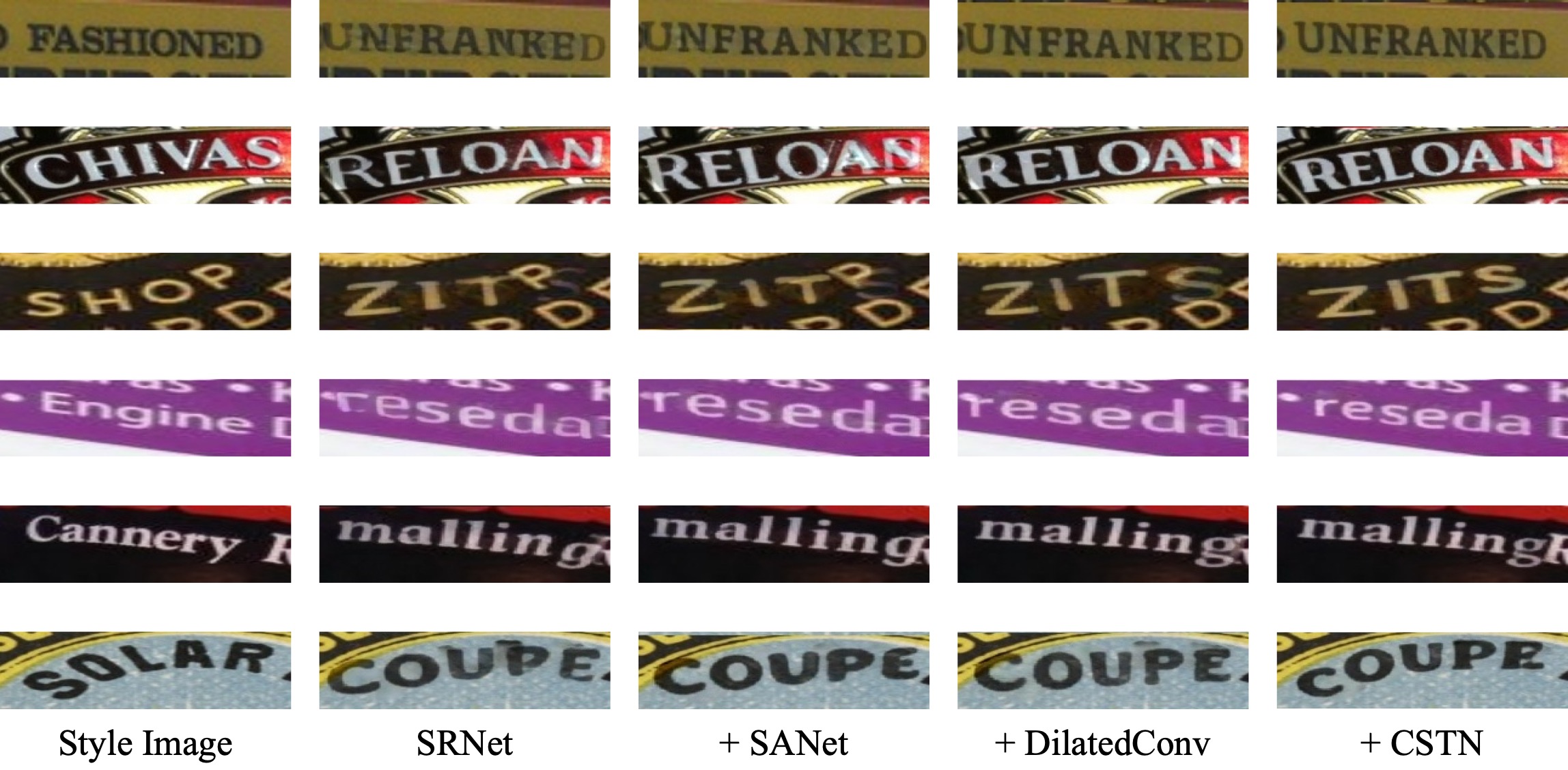}
	\caption{Some results of ablation study.}
	\label{fig:ablation}
\end{figure}

In this section, we empirically investigate how the performance of our proposed framework is affected by different model settings.
Our study mainly focuses on these aspects: the content shape transformation network, the self-attention network, and dilated convolution in background completion network.
Some qualitative results are presented in Figure~\ref{fig:ablation} 

\paragraph{Content Shape Transformation Network (CSTN)}

Content shape transformation network (CSTN) aims to transform the content image according to the geometrical attributes of the style image.
This is critical for text style transfer in real-world images, because scene text images often contain severe geometric distortion, such as in arbitrary oriented, perspective or curved form.
With CSTN, the coherence of geometrical attributes between content and style images could be achieved.
Although the whole model is difficult to train on real images, the CSTN can be finetuned on real datasets. As illustrated
in Figure ~\ref{fig:ablation}, the positions of generated text are more plausible.
Quantitative results of CSTN is shown in Table~\ref{table:ablation}, the PSNR increased by over $0.35$ and SSIM increased by over $0.017$ on average.

\paragraph{Self-attention Network}

Self-attention network is used to adequately combine the content features and style features.
According to Table~\ref{table:ablation}, with single level self-attention network, the average $l_2$ error is decreased by about $0.003$, the average PSNR is increased by about $0.3$, and the average SSIM is increased by about $0.012$.
To use more global statistics of the style and content features, we adopt a multi-level self-attention network to fuse global and local patterns.
With multi-level self-attention network, all the metrics have been improved.

\paragraph{Dilated Convolution}

Dilated convolutional layers can enlarge the pixel regions to reconstruct the background images, therefore, it is easier to generate higher quality images.
According to Table~\ref{table:ablation}, the background completion network with dilated convolutional layers has a better performance on all metrics.

\begin{table}[h]
	\centering
	\fontsize{6.5}{6.5}\selectfont
	\setlength{\tabcolsep}{6pt}
	\renewcommand{\arraystretch}{2.0}
	\caption{Quantitative results on synthetic test dataset.}
	\begin{tabular}{|l|c|c|c|c|c|c|}
		\hline
		\multirow{2}*{Method} & \multicolumn{3}{|c|}{English} & \multicolumn{3}{|c|}{Chinese}  \\ \cline{2-7}
		~ & $l_2$ & PSNR & SSIM & $l_{2}$ & PSNR & SSIM  \\ \hline
		pix2pix~\cite{isola2016image-to-image} & 0.0953 &	12.32 &	0.551 & 0.11531 & 10.09 & 0.3523 \\ 
		SRNet~\cite{Wu:2019ww} & 0.0472 &	14.91&	0.6213&	0.0512&	14.77&	0.5719 \\ \hline
		w/o CSTN &	0.0436	& 15.22	& 0.6375&	0.0463&	14.98 & 0.5903 \\
		w/o SA &	0.0421 &	15.31 &	0.6401 &	0.0459 &	15.02 &	0.5987 \\
		w/o DilatedConv & 0.0402 &	15.23 &	0.6479 &	0.0432 & 	15.15 &	0.6032 \\ \hline
		SwapText (single) & 0.0397 &	15.53 &	0.6523 &	0.0422 &	15.38 &	0.6112 \\
		SwapText (multi) & \bf0.0381 &	\bf16.04 &	\bf0.6621	& \bf0.0420 &	\bf15.46 &	\bf0.6189 \\
		\hline
	\end{tabular}
	
	\label{table:ablation}
\end{table}

\subsection{Comparison with Prior Work}
To evaluate our proposed method, we compared it with two types of text swapping method: pix2pix proposed in~\cite{isola2016image-to-image} and SRNet proposed by Wu \etal.~\cite{Wu:2019ww}.
We use our generated datasets to train and test these two models.
Both methods maintain the same configurations according to the papers.

\paragraph{Quantitative results}
In Table~\ref{table:ablation}, we give some quantitative results of our method and other two competing methods.
Clearly, our proposed method has a significant improvement on all the metrics across different languages.
The average $l_2$ error is decreased by over $0.009$, the average PSNR is increased by over $0.9$, and the average SSIM is increased by over $0.04$ than the second best method.

To further evaluate the quality of generated images, we propose to use text recognition and detection accuracy on generated images.
We use the text recognition model CRNN to evaluate our generated images on SVT-P, IC13 and IC15 dataset.
The CRNN model is trained on the mix of training images on these datasets and the recognition accuracy is present in Table~\ref{table:recog}. 
On IC13, the recognition accuracy is even higher than the real test set.
We use an adapted version of EAST~\cite{zhou2017east} to detect text in the images. 
Since the implementation of the original EAST is not available, we use the public implementation\footnote{\url{https://github.com/argman/EAST}} with ResNet-50 backbone.
We replace the texts in images of IC13 and IC15 test sets, then evaluate the generated datasets using the model trained on IC13 and IC15 training datasets.
According to the comparison results presented in Table~\ref{table:detection}, the F-measure on generated IC13 and IC15 test sets are $78.4\%$ and $80.2\%$ respectively, which is close to the metrics on real test set.
This indicates that the images generated by our framework are very realistic and can even fool the text detection model.

\begin{table}[th]
	\centering
	\small
	\fontsize{8}{8}\selectfont
	\setlength{\tabcolsep}{8pt}
	\renewcommand{\arraystretch}{2.0}
	\caption{Comparison of text recognition accuracy on real and generated images.}
	\begin{tabular}{|c|c|c|c|}
		\hline
		Dateset & SVT-P & IC13 & IC15 \\ \hline
		Real & \bf54.3 & 68.0 & \bf 55.2 \\ \hline
		pix2pix & 22.1 & 34.7 & 25.8 \\ \hline
		SRNet & 48.7 & 66.8 & 50.2 \\ \hline
		Generated & 54.1 &  \bf 68.3 & 54.9 \\ \hline
		
	\end{tabular}
	
	\label{table:recog}
\end{table}

\begin{table}[th]
	\centering
	\fontsize{8}{8}\selectfont
	\setlength{\tabcolsep}{8pt}
	\renewcommand{\arraystretch}{2.0}
	\caption{Comparison of text detection accuracy between real data and generated data on IC13 and IC15 datasets.}
	\begin{tabular}{|c|c|c|c|c|c|c|}
		\hline
		\multirow{2}*{Test Set} & \multicolumn{3}{|c|}{IC13} & \multicolumn{3}{|c|}{IC15} \\ \cline{2-7}
		~ & R & P & F & R & P & F \\ \hline
		Real & 74.5 & 84.0 & 79.0 & 77.3 & 84.6 & 80.8 \\ \hline
		pix2pix & 66.4 & 80.7 & 72.8 & 71.8 & 79.3 & 75.3 \\ \hline
		SRNet & 70.4 & 82.9 & 76.1 & 74.2 & 82.5 & 78.1 \\ \hline
		SwapText &  73.9 & 83.5 & 78.4 & 76.8 & 84.1 & 80.2 \\ \hline
	\end{tabular}
	
	\label{table:detection}
\end{table}

\subsection{Image Based Text Translation}

\begin{figure}[h]
	\centering
	\includegraphics[width=0.35\textwidth]{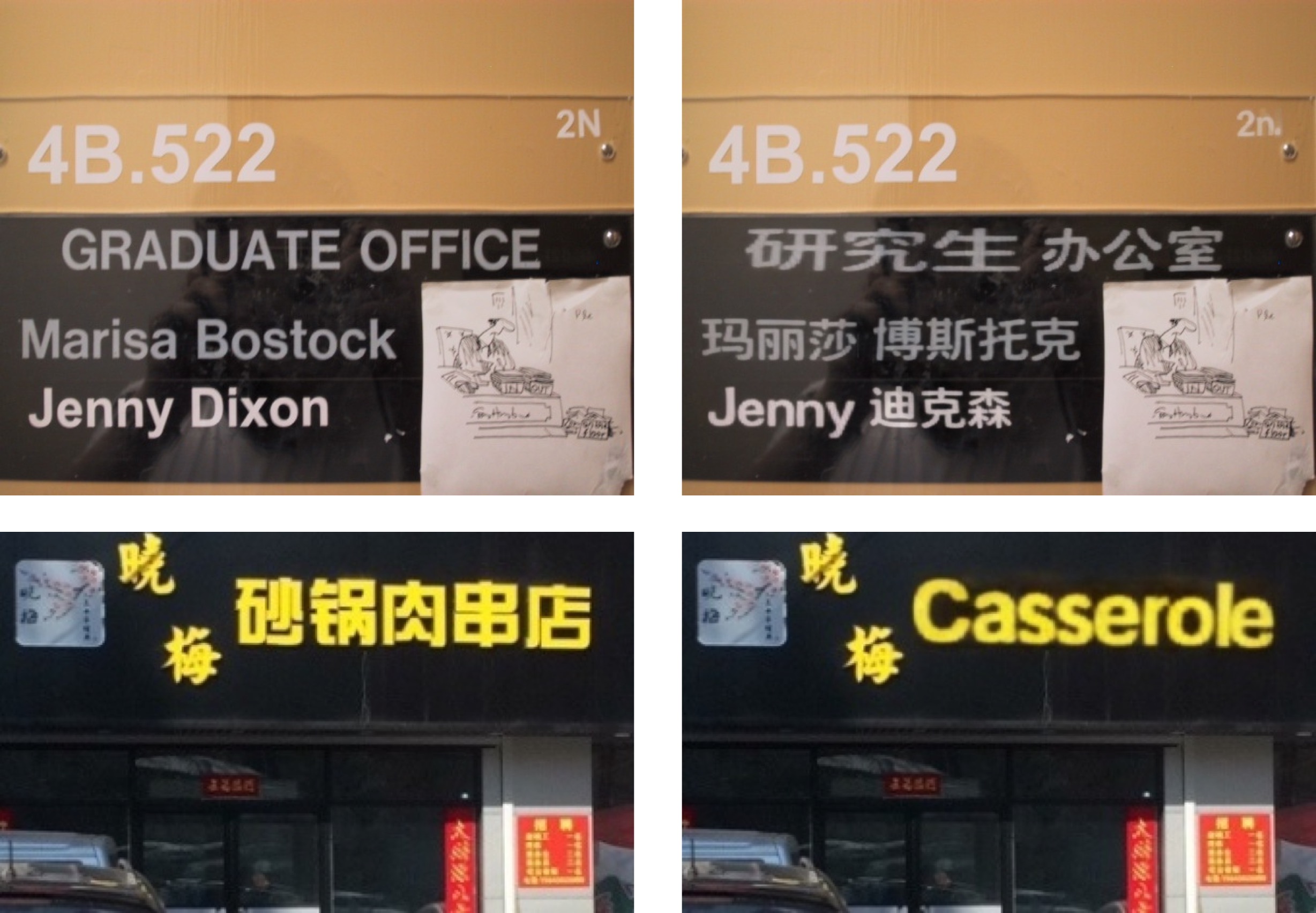}
	\caption{Image based translation examples. (\textit{Left}) Input images. (\textit{Right}) Translation results.}
	\label{fig:trans}
\end{figure}

Image based translation is one of the most important applications of arbitrary text style transfer.
In this section, we present some image based translation examples, which are illustrated in Figure~\ref{fig:trans}. 
We conduct translation between english and chinese.
According to the results, we can find that no matter the target language is chinese or english, the color, geometric deformation and background texture can be kept very well, and the structure of characters is the same as the input text.

\begin{figure}[th]
	\centering
	\includegraphics[width=0.3\textwidth]{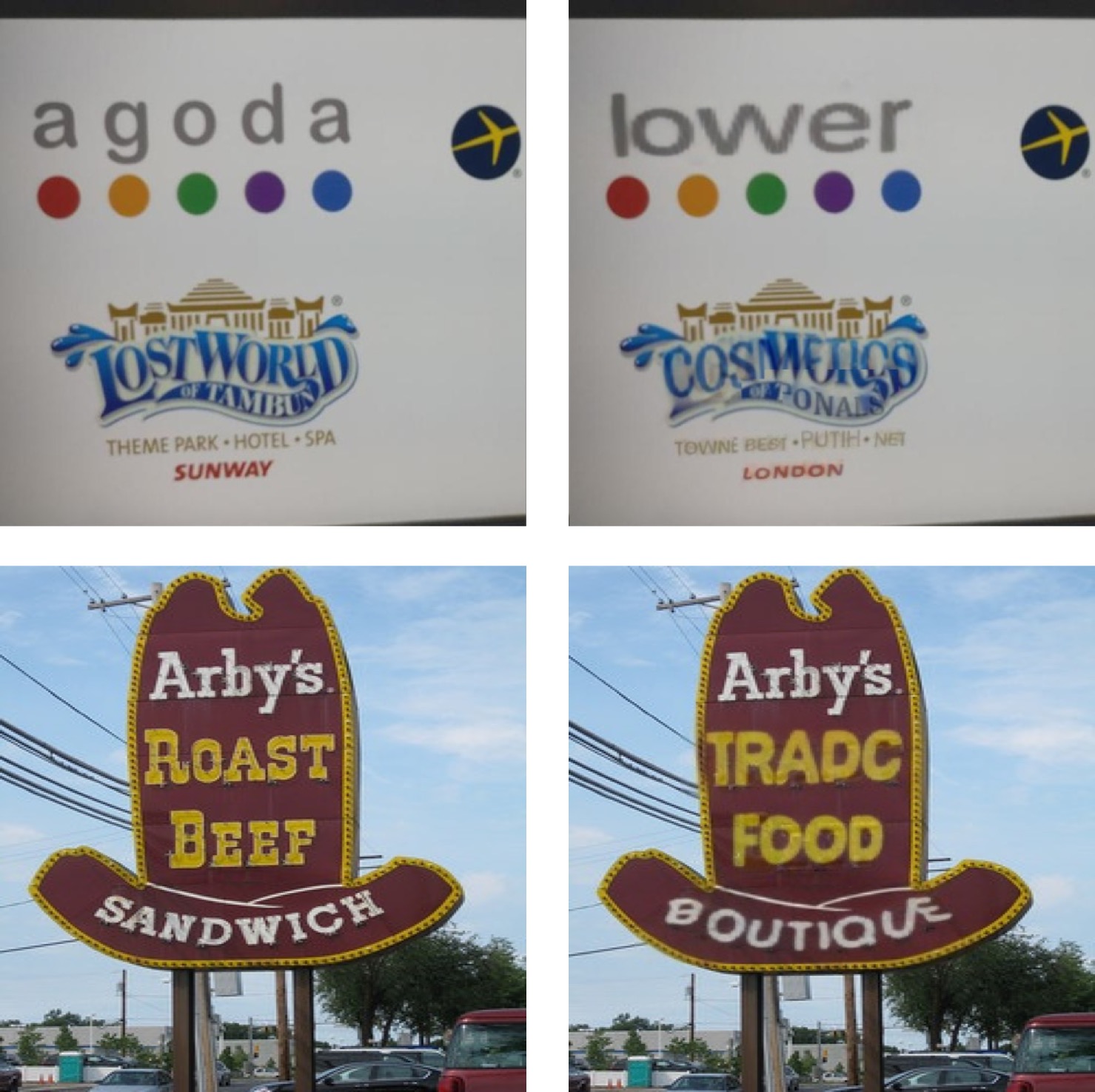}
	\caption{Failure Cases. (\textit{Top}) Wavy text. (\textit{Bottom}) WordArt.}
	\label{fig:fail}
\end{figure}

\begin{figure*}[th]
	\centering
	\subfigure[Generated images on IC15 dataset.]{
		\includegraphics[width=0.85\textwidth, height=0.35\textheight]{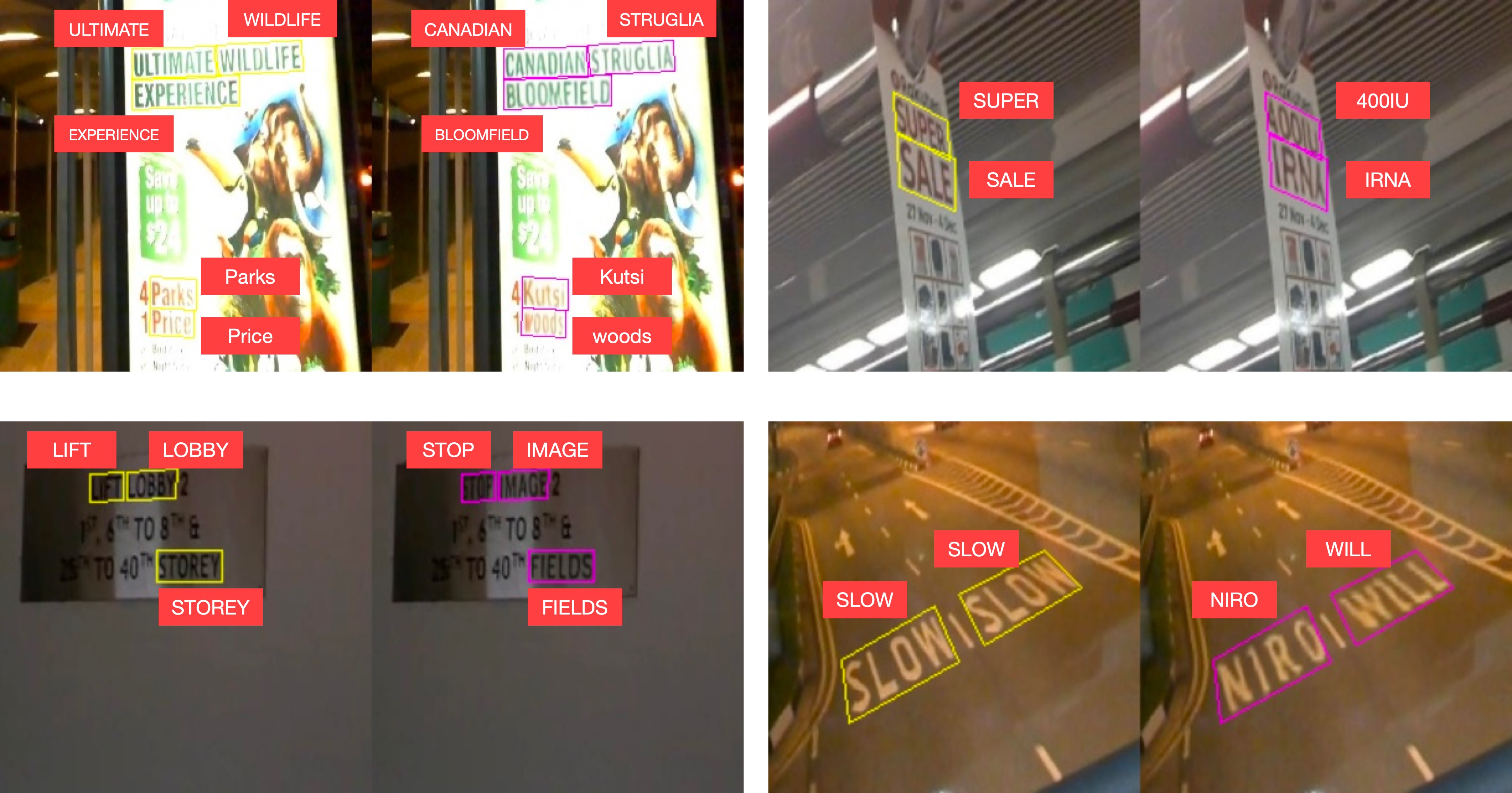}
	}
	\subfigure[Generated images on IC17 dataset.]{
		\includegraphics[width=0.85\textwidth, height=0.35\textheight]{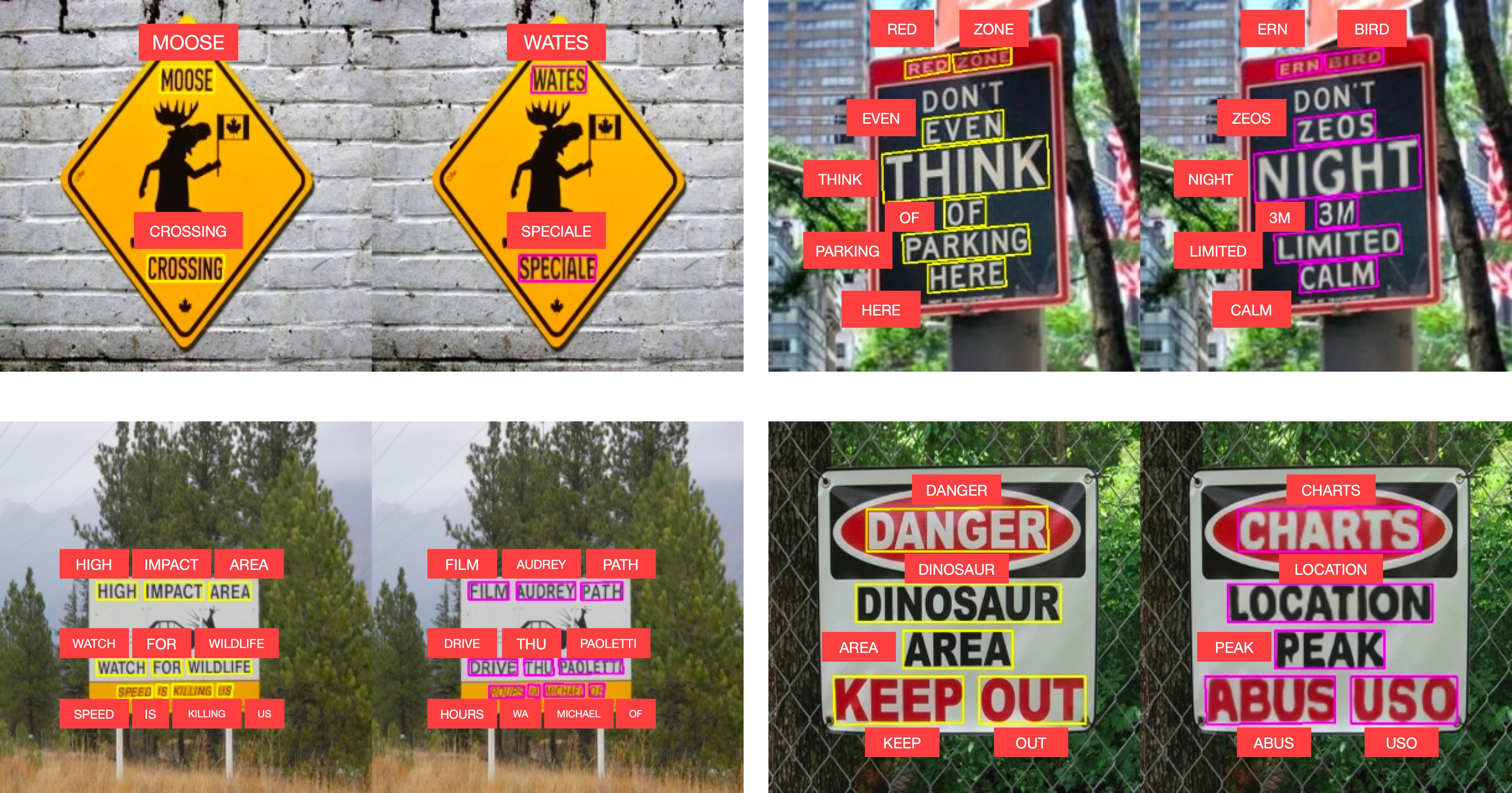}
	}
	\caption{Generated images on scene text datasets. The image on the left is the original image, while the right one is the generated image. }
	
	\label{fig:samples}
\end{figure*}

In Figure~\ref{fig:samples}, we also present some example results of our model evaluated on scene text datasets.  
According to Figure~\ref{fig:samples}, our model can replace the text in the input image while keeping the original fonts, colors, sizes and background textures.

\subsection{Limitations}
Our method has the following limitations.
Due to the limited amount of training data, the geometric attribute space and font space are not fully exploited.
Our proposed method fails when the text in style image is waved, see Figure~\ref{fig:fail} (Top).
Figure~\ref{fig:fail} (Bottom) shows a failure case on style image with WordArt.

\section{Conclusion}

In this study, we proposed a robust scene text swapping framework SwapText to address a novel task of replacing texts in the scene text images by intended texts.
We adopt a divide-and-conquer strategy, decompose the problem into three sub-networks, namely text swapping network, background completion network and the fusion network.
In the text swapping network, the features of content image and style image are extracted simultaneously and then combined by a self-attention network.
To better learn the representation of content image, we use a Content Shape Transformation Network (CSTN) to transform the content image according to the geometrical attributes of the style image.
Then, a background completion network is used to generate the background image of style image.
Finally, the output of text swapping network and background completion network are fed into the fusion network to generate more realistic and semantically coherent images.
Qualitative and quantitative results on several public scene text datasets demonstrate the superiority of our approach. 

In the future work, we will explore to generate more controllable text images based on the fonts and colors.

\newpage
{\small
	\bibliographystyle{ieee_fullname}
	\bibliography{main}
}

\end{document}